\documentclass[11pt, a4paper, logo, copyright, nonumbering]{metastone}
\usepackage[authoryear, sort&compress, round]{natbib}
\usepackage{dblfloatfix}
\usepackage{ulem}
\usepackage{caption}
\usepackage{dramatist}
\usepackage{xspace}
\usepackage{pifont} %
\usepackage{multirow}
\usepackage{tcolorbox}
\usepackage{xltabular}
\usepackage{longtable}
\usepackage{hyperref}
\usepackage{amsfonts}
\usepackage{amsmath}
\usepackage{amssymb}
\usepackage{lineno}
\usepackage{multirow}
\usepackage{adjustbox}

\usepackage[bottom]{footmisc}

\usepackage{CJKutf8}
\usepackage{subfigure}
\usepackage{setspace}

\usepackage{dsfont}
\usepackage{array} %
\usepackage{tabularx} %
\usepackage{subfigure} %
\usepackage{xcolor} %
\usepackage{tabularx}
\usepackage{booktabs}

\usepackage{lipsum}  %
\usepackage{multicol} %

\usepackage{algorithm}
\usepackage{algorithmic}

\newmdenv[
  tikzsetting={draw=red, line width=1.2pt, dashed},
  leftline=true,
  rightline=true,
  topline=true,
  bottomline=true,
  innertopmargin=6pt,
  innerbottommargin=6pt,
  innerleftmargin=6pt,
  innerrightmargin=6pt,
  backgroundcolor=white
]{mydashedbox}

\makeatletter
\def\@BTrule[#1]{%
  \ifx\longtable\undefined
    \let\@BTswitch\@BTnormal
  \else\ifx\hline\LT@hline
    \nobreak
    \let\@BTswitch\@BLTrule
  \else
     \let\@BTswitch\@BTnormal
  \fi\fi
  \global\@thisrulewidth=#1\relax
  \ifnum\@thisruleclass=\tw@\vskip\@aboverulesep\else
  \ifnum\@lastruleclass=\z@\vskip\@aboverulesep\else
  \ifnum\@lastruleclass=\@ne\vskip\doublerulesep\fi\fi\fi
  \@BTswitch}
\makeatother

\addto\extrasenglish{
}

 {\begin{list}{}%
         {\setlength{\leftmargin}{#1}}%
         \item[]%
 }
 {\end{list}}
 
\reportnumber{001} %

\renewcommand{\today}{}

\title{\centering Test-Time Scaling with Reflective Generative Model}

\author[*]{
Zixiao Wang\textsuperscript{2*}, Yuxin Wang\textsuperscript{1*}, Xiaorui Wang\textsuperscript{1}, Mengting Xing\textsuperscript{1}, Jie Gao\textsuperscript{1}, Jianjun Xu\textsuperscript{2}, Guangcan Liu\textsuperscript{1}, Chenhui Jin\textsuperscript{2}, Zhuo Wang\textsuperscript{2}, Shengzhuo Zhang\textsuperscript{1}, Hongtao Xie\textsuperscript{2$\dagger$}
\small
\\
MetaStone-AI\textsuperscript{1} \& USTC\textsuperscript{2}
}

\renewcommand{\phi}{\varphi}

\renewcommand{\epsilon}{\varepsilon}
\renewcommand{\imath}{\mathrm{i}}

\newlength{\restsubwidth}
\newlength{\restsubheight}
\newlength{\restsubmoreheight}
\newcommand{\rest}[2]{%
        \settowidth{\restsubwidth}{\ensuremath{#2}}
        \settoheight{\restsubheight}{\ensuremath{{}_{#2}}}
        \ensuremath{{#1\hskip 0.5pt}_{\vrule\kern2pt\parbox[b][%
        4pt][b]{\the\restsubwidth}{%
                        \ensuremath{{}_{#2}}}}}
        }

\footnotetext{*Equal Contribution. $\dagger$ Corresponding author.}

\begin{abstract}

We introduce our first reflective generative model MetaStone-S1, which obtains OpenAI o3-mini's performance via the new Reflective Generative Form. The new form focuses on high-quality reasoning trajectory selection and contains two novelties: 1) \textbf{A unified interface for policy and process reward model}: we share the backbone network and use task-specific heads for reasoning trajectory predicting and scoring respectively, introducing only 53M extra parameters for trajectory scoring. 2) \textbf{Eliminating the reliance on process-level annotation}: we provide a self-supervised process reward model, which can directly learn the high-quality reasoning trajectory selection from the outcome reward. Equipped with the reflective generative form, MetaStone-S1 is naturally suitable for test-time scaling, and we provide three reasoning effort modes (low, medium, and high) based on the controllable thinking length. Experiments demonstrate that our MetaStone-S1 achieves comparable performance to OpenAI o3-mini's series with only 32B parameter size. To support the research community, we have open-sourced MetaStone-S1 at \url{https://github.com/MetaStone-AI/MetaStone-S1}.
\end{abstract}

\begin{document}

\maketitle

\begin{figure}[!h]
\centering
\includegraphics[width=0.9\textwidth]{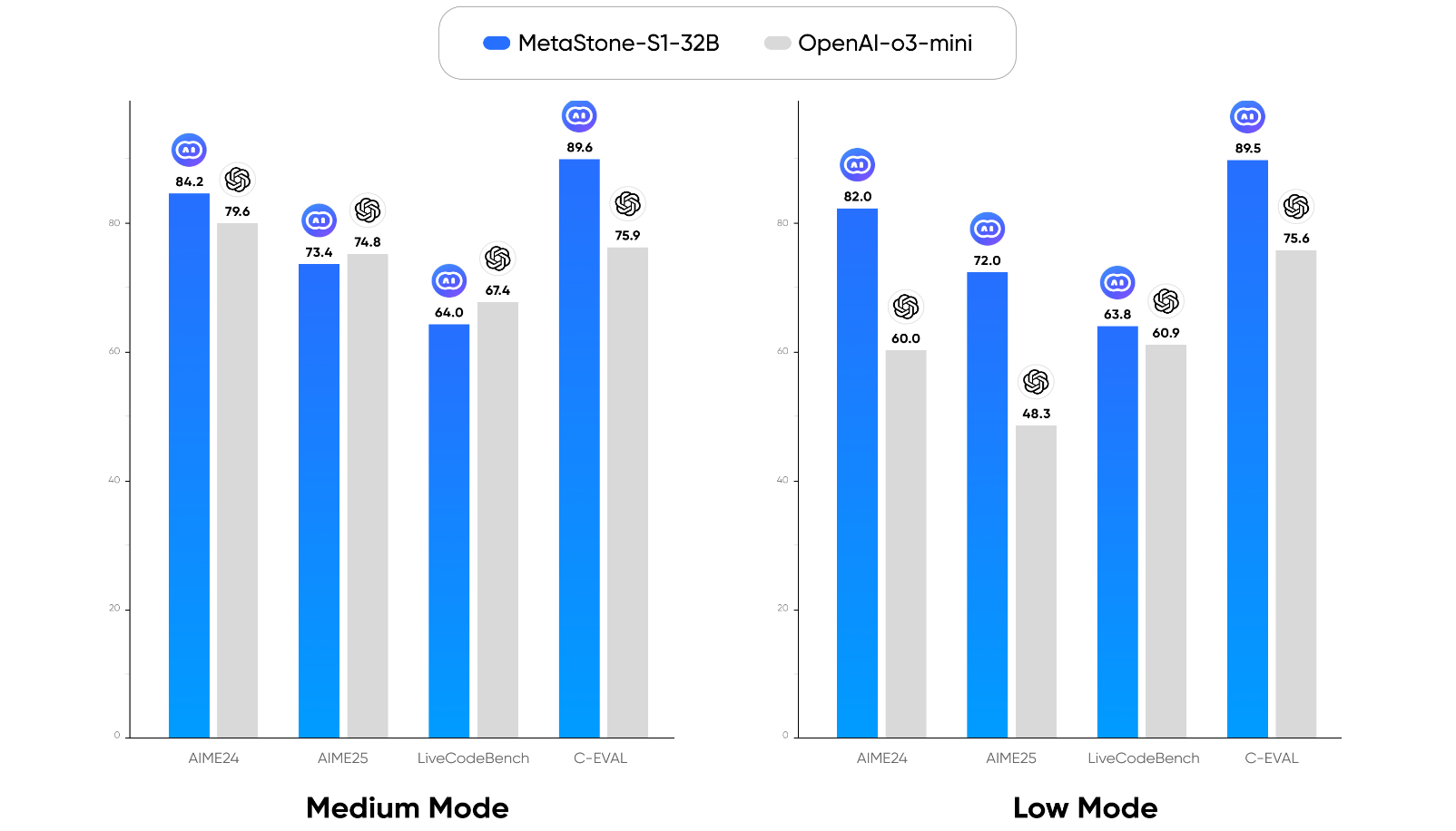}
\vspace{-5pt}
\caption{
    \centering
    Benchmark performance of MetaStone-S1.
}
\label{fig:performance}
\end{figure}


\newpage

\section{Introduction}
Over the past two years, the field of Large Language Models (LLMs) has experienced rapid advancements, marked by the emergence of increasingly sophisticated models. Notable developments include OpenAI’s GPT-4, Google’s Gemini, Meta’s LLaMA series, Alibaba's Qwen, and DeepSeek’s R1, which have collectively pushed the boundaries of natural language understanding and generation.  This progress is attributed to innovations in model architectures and training techniques, enabling LLMs to process and generate content across various formats. 

Recent analyses suggest that OpenAI’s o3 model achieves its advanced reasoning and coding capabilities through Test-Time Scaling (TTS) techniques such as massive sampling, candidate scoring, and search over multiple reasoning paths \citep{adaline2024o3,analysis2024o3}.
For instance, during ARC-AGI and competitive coding evaluations, o3 was shown to generate up to 1024 candidate samples for each query \citep{2024arc,openai2024o3}.
These inference-time strategies mark a significant shift from traditional one-pass models, enabling o3 to adapt dynamically to novel tasks and achieve near-human performance in reasoning benchmarks.

TTS approaches can be categorized into two types: internal TTS and external TTS. Internal TTS (also called sequential TTS in \cite{zeng2025revisiting}) strategies use CoT for longer thinking processes \citep{deepseekr1,openai2024o1}, which benefits from Long-CoT Supervised Fine-Tuning and reinforcement learning. 
Recent internal TTS methods \citep{deepseekr1} mainly suffer from the false positive reasoning process, as the outcome reward will misclassify the correct answer with incorrect reasoning during the training stage. External TTS (also called parallel TTS in \cite{zeng2025revisiting}) is proposed for selecting the correct reasoning process, which is proved to be more effective in performance boosting compared with outcome reward \citep{lightman2023let}. Prominent external TTS algorithms include Best-of-N sampling, Beam Search, and Diverse Verifier Tree Search, using the verifier such as the Process Reward Model(PRM) to select high-quality reasoning trajectories. For example, Best-of-N firstly generates multiple outputs, and then sequentially uses PRM-based to score and select the best solution. \cite{liu2025can} point out that the best external TTS solution varies from policy model with different parameters. When the parameter of the policy model is less than 32B, the search methods (Beam Search and Diverse Verifier Tree Search) achieve better results than the sampling one(Best-of-N). However, when the parameter size is equal to or greater than 32B, the sampling method can achieve better performance. Based on the above analysis, internal and external TTS are two individual methods and can benefit each other, e.g. using the Best-of-N to boost the Long-CoT model with PRM.

This paper focuses on external TTS and proposes a new Reflective Generative Form for high-quality reasoning trajectory selection. Specially, the proposed new form shares the backbone of the policy model and process reward model, and uses self-supervised training to eliminate the reliance on process-level supervision. Based on the Reflective Generative Form, the proposed MetaStone-S1 contains high, medium, and low reasoning modes with the controllable thinking length. Experiment results show that MetaStone-S1 achieves comparable performance to OpenAI o3-mini's series with only 32B parameters.

\subsection{Contributions}

\begin{itemize}[topsep=0pt]
    \item \textbf{A new Reflective Generative Form}: We systematically review the existing Test-Time Scaling (TTS) paradigms, and provide a clear definition of the Reflective Generative Form for high-quality reasoning trajectory selection. The proposed Reflective Generative Form enables a single network to achieve both reasoning trajectory prediction and selection (with \textbf{Zero} process-level annotation).
    \item \textbf{The comprehensive analysis about the aha moment, scaling law and robustness of the new form}: We provide both qualitative and quantitative analysis for the aha moment, scaling law, and robustness of the proposed new form. These exhaustive discussions will effectively benefit the community for future research.
    \item \textbf{The State-of-the-art performance}: MetaStone-S1 achieves comparable performance as the OpenAI o3-mini’s series with only 32B parameter. Specially, MetaStone-S1-low outperforms OpenAI o3-mini-low in mathematical(AIME24\&25), coding(LiveCodeBench) and Chinese reasoning(C-Eval) tasks respectively. MetaStone-S1-medium obtains similar results to OpenAI o3-mini-medium. Finally, MetaStone-S1-high further improves the intelligence ceiling and achieves SOTA results in a series of open-source and closed-source models.
\end{itemize}

\section{Related Works}
\subsection{Test-Time Scaling}
Test-Time Scaling (TTS) is a technique that leverages additional computational resources at inference time to tackle challenging problems. 
With the remarkable performance improvements demonstrated by OpenAI o1 \citep{jaech2024openai}, TTS has become a research hot spot for enhancing the reasoning capabilities of LLMs.
TTS can be divided into two categories: internal TTS and external TTS.
Internal TTS introduces the long Chain-of-Thought (CoT) to generate answers based on the detailed reasoning process.
OpenAI o1\citep{jaech2024openai} and DeepSeek R1\citep{deepseekr1} introduce a thinking process to plan the solution and guide the final answer.
\cite{jin2024impact,yeo2025demystifying} have shown that long CoT can help models correct mistakes by themselves and decompose complex problems more effectively, thereby improving performance.
DeepScaleR\citep{deepscaler2025} demonstrates that by carefully extending the context length during training, only a 1.5B-parameter model can surpass the o1-Preview.
However, \cite{chen2024not,chen2024unlocking} have highlighted the risk of overthinking, where excessively long reasoning trajectories may lead to performance degradation.
On the other hand, external TTS scales up inference through search-based strategies and auxiliary reward models.
A common approach is the Best-of-N strategy \citep{lightman2023let,brown2024large,wang2023math}, which generates multiple candidates and selects the best one based on scores from the pretrained reward model.
Moreover,  fine-grained methods have also been explored, including, such as Beam Search \citep{liu2025can,snell2024scaling}, Diverse Verifier Tree Search \citep{beeching2024scalingtesttimecompute} and Monte Carlo Tree Search (MCTS) \citep{zhang2024rest,guan2025rstar,luo2024improve}.
These methods search at the step level and utilize Process Reward Models (PRMs) to guide the reasoning trajectory step-by-step.
Beyond search strategies, recent work emphasizes that the quality of the reward model is a crucial factor in external TTS \citep{guan2025rstar}.
A straightforward and effective way to enhance a model’s reasoning ability is to develop a high-quality reward model.

\subsection{Process Reward Model}
Process Reward Models (PRMs) focus on evaluating LLMs at the step level.
\cite{lightman2023let} unveil that this fine-grained guidance can lead to better TTS performance compared with the global-level Outcome Reward Model (ORM).
However, accurately identifying logical errors in LLM outputs remains challenging, and PRMs require high-quality task-specific annotated data for training.
To this end, recent works \cite{wang2023math} leverage Monte Carlo estimation to automatically assign step-level scores using only the final answers as supervision.
\cite{zhang2024rest, guan2025rstar} iteratively synthesizes data by MCTS and fine-tuning both LLMs and PRMs, improving performance across both models.
\cite{tan2025aurora} follow the LLM-as-a-judge method and introduce a new LLM to annotate the reward of each step.
Nonetheless, \cite{zhang2025lessons} point out that labels generated by Monte Carlo estimation can be noisy, as incorrect reasoning processes may still yield correct final answers. 
They further propose a hybrid approach that combines both Monte Carlo estimation with the LLM-as-a-judge.

Despite these advances, existing PRMs still suffer from several challenges.
First, the PRMs are trained on a new large-scale LLM model, resulting in significant training and inference costs.
Second, most PRM training methods typically follow an off-policy strategy, which limits their ability to directly discriminate outputs generated by the target LLM. 
The unseen distribution during inference time may further degrade performance.
To address these issues, we propose a Reflective Generative Form, which shares most parameters between the PRM and the target LLM, and supports on-policy optimization with only outcome rewards, enabling more efficient and aligned training.

\section{Problem Formulation}

This paper aims to find a high-quality reasoning trajectory more efficiently at inference time based on TTS.
We first summarize the general inference forms for standard LLMs (policy models) and existing TTS methods, and then formally define our proposed Reflective Generative Form.

\noindent \textbf{1) Basic LLMs.} The model directly generates an answer based on the input query.
This basic inference form can be formulated as:
\begin{equation} \label{eq:basic_form}
\text{answer} = LLM_{\text{answer}}(\text{query}).
\end{equation}

TTS based methods can be categorized into two types: sequential scaling based internal TTS and parallel scaling based external TTS.

\noindent \textbf{2) Internal TTS (e.g. DeepSeek R1\citep{deepseekr1}).} The internal TTS first generates a reasoning trajectory by Long-CoT using $LLM_{\text{thinking}}$, and then predicts the final answer based on this trajectory using $LLM_{\text{answer}}$.
This procedure can be expressed as:
\begin{equation} \label{eq:internaltts_form}
\text{answer} = LLM_{\text{answer}}(LLM_{\text{thinking}}(\text{query})).
\end{equation}
To be specific, recent methods (e.g. DeepSeek R1\citep{deepseekr1}) use the same policy model for both $LLM_{\text{thinking}}$ and $LLM_{\text{answer}}$.

\noindent \textbf{3) External TTS (e.g. \cite{lightman2023let,liu2025can,zhang2024rest}).} Firstly, the Long-CoT generation is extended by generating multiple reasoning trajectories and answers in parallel. Then, a reward model (e.g. PRM) is used to score and select the best result.
This inference form can be described as:
\begin{equation} \label{eq:externaltts_form}
\text{answer} = \underset{i \in [1,k]}{\arg\max} \; LLM_{PRM}\Big( [LLM_{\text{answer}}(LLM_{\text{thinking}}(\text{query}))]_i \Big),
\end{equation}
where $[*]_i$ denotes the $i$-th candidate among $k$ parallel generations.

Though existing external TTS methods have been proven to obtain considerable performance enhancement, they still encounter several problems:
(1) Extra Computation: PRM contains individual parameters from the policy model($LLM_{\text{think}}$ and $LLM_{\text{answer}}$), which introduces additional huge computation. 
(2) Expensive Annotation: It is difficult to obtain the large-scale reasoning trajectory annotations for PRM training.

\noindent \textbf{Reflective Generative Form.} To address the extra computation and expensive annotation issues, we propose a new Reflective Generative Form focusing on the efficient and label-free reasoning trajectory selection. The proposed Reflective Generative Form is shown in Eq.\ref{eq:s1_form}. 

\begin{equation} \label{eq:s1_form}
\text{answer} = 
\underbrace{LLM_{\text{answer}}}_{\text{share backbone}}
\Big(
  \underset{i \in [1,k]}{\arg\max} \;
  \underbrace{LLM_{SPRM}}_{\text{share backbone}}
  \Big( [\underbrace{LLM_{\text{thinking}}}_{\text{share backbone}}(\text{query})]_i \Big)
\Big)
\end{equation}

Firstly, we share the backbone of the policy model and PRM in a single network, which enables reasoning trajectory generation and scoring in a unified interface for parallel prediction.
The score measures the quality of each reasoning trajectory, and the trajectory with higher score is selected as the high-quality candidate in TTS.
This unified interface is proved to be effective for parameter reduction in our experiments.
Secondly, we introduce a novel Self-supervised Process Reward Model (SPRM) to eliminate the reliance on process-level annotation, which can be optimized with only outcome-level annotation in a self-supervised manner. 
In particular, we only implement the SPRM for the $LLM_{\text{think}}$ selection, which can further improve the inference efficiency during the real implementation.

\section{Approach}

\begin{figure*}[!t] 
  \centering
  \includegraphics[width=0.8\textwidth]{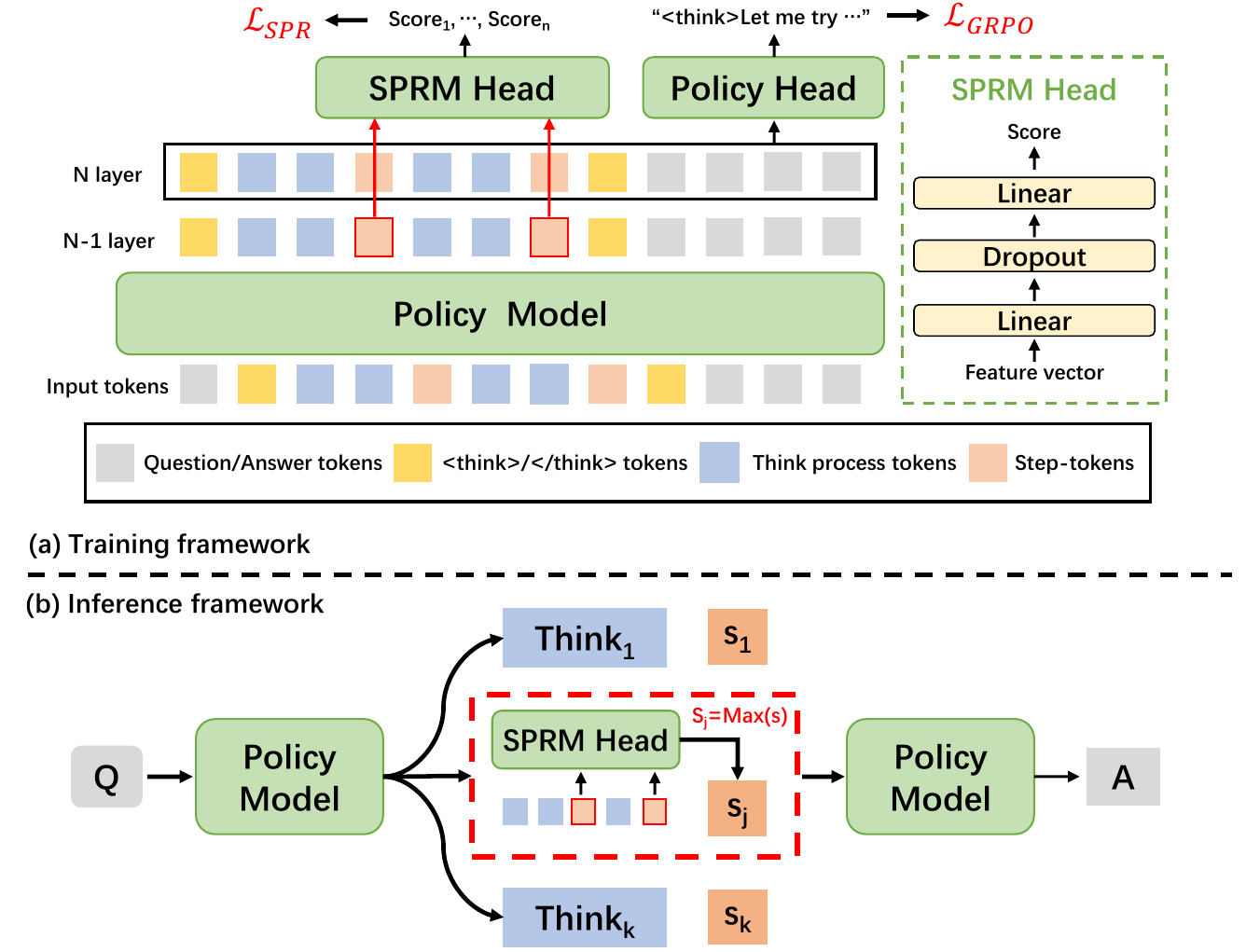}
  \caption{The training and inference framework of Reflective Generative Models.}
  \label{fig:img_framework}
\end{figure*}

\subsection{Unified Interface in Reflective Generative Form}
Our proposed Reflective Generative Form establishes a unified interface for the policy model and the PRM.
For the policy model, we employ reasoning LLMs that contain the thinking process in response, delineated by the '<think>' and '</think>' tokens.
For the PRM, we introduce a Self-supervised Process Reward Model (SPRM), which shares the same backbone as the policy model but incorporates an additional lightweight SPRM head.
The SPRM head is implemented by a binary classifier consisting of two linear layers and a dropout layer.
An overview of the joint framework is illustrated in Fig.~\ref{fig:img_framework}(a).

Within this unified form, the policy model first generates multiple thinking processes as the reasoning trajectories.
Subsequently, the SPRM evaluates each thinking process for reasoning trajectory selection.
The evaluation procedure contains two steps: (1) Segmenting the reasoning trajectory into discrete steps and (2) Predicting a trajectory score based on evaluation in each step.

\noindent \textbf{1. Step Segmentation.}
We segment each reasoning trajectory using tokens that are already supported by the policy model’s tokenizer, eliminating the need to introduce additional step-specific tokens or fine-tune the LLM for step-format outputs.
Specifically, we treat tokens containing '.\verb|\n\n|' as step-tokens and split the trajectory accordingly.
Additionally, we retain only the first token in any sequence of consecutive step-tokens and ignore the step-token appearing at the beginning of the trajectory, as it does not contain substantive solution information.

\noindent \textbf{2. Trajectory Score Prediction.}
After using step-tokens to mark the end of individual reasoning steps, we evaluate each step based on the representation of the corresponding step-token.
Since the representation in the last layer mainly captures the logits prediction for a single token, we use the hidden representations from the second-to-last layer of the policy model to provide richer contextual information of the entire step.
These representations are then fed into the SPRM head to predict process scores for each step.
The final score for the entire reasoning trajectory is computed as the geometric mean of the individual process scores:
\begin{equation} \label{eq:final_score}
\text{S}_{\text{final}} = \left( \prod_{i=1}^{n} \text{Score}_i \right)^{\frac{1}{n}} = \left( \prod_{i=1}^{n} SPRM(f_{token_i}) \right)^{\frac{1}{n}},
\end{equation} 
where $n$ denotes the total number of steps, and $f_{\text{token}_i}$ is the representation of the $i$-th step-token obtained from the policy model. $\text{Score}_i$ is the SPRM's process score for $i$-th step.

Through this unified interface, a single network can generate reasoning trajectories and score them in parallel, enabling joint training in an end-to-end manner.
This design facilitates a straightforward and efficient training pipeline for on-policy PRM learning, where both the policy model and the SPRM continuously refine their parameters from shared experiences, thereby improving the overall quality of the generated trajectories.

\subsection{Optimization of Reflective Generative Form}
During optimization, we train the policy model and the SPRM head simultaneously.
For optimizing the policy model, we adopt Group Relative Policy Optimization (GRPO) following \cite{shao2024deepseekmath}.
To optimize the SPRM head, we propose a Self-supervised Process Reward Loss (SPR Loss), which enables learning process discrimination ability only from outcome reward (e.g. final answer correctness).
The SPR Loss is formulated as follows, 
\begin{equation} \label{eq:sprloss}
\mathcal{L}_{\text{SPR}} = \frac{1}{N}\sum_{i=1}^{N} w_i * BCELoss(\text{Score}_{i},y_i), \quad
\text{where} \ w_{i} = 
\begin{cases}
1, & \text{if} \ y_i=1 \ \& \ \text{Score}_{i}>0.5 \\ 
1, & \text{if} \ y_i=0 \ \& \ \text{Score}_{i}<0.5 \\ 
0, & \text{others} 
\end{cases}
,
\end{equation}
where $i$ denotes the step-tokens, $w_{i}$ is a token-level weight, $\text{Score}_{i}$ is SPRM's process score on step $i$, and $y_{i}$ denotes whether the final answer from the policy model is correct.
In Eq.\ref{eq:sprloss}, the process score is optimized based on the correctness of the final answer.
However, since a correct final answer may include incorrect intermediate steps and vice versa \citep{lightman2023let}, we introduce the self-supervised dynamic weight $w_i$ to mitigate supervision noise.
Specifically, we use the SPRM head's own prediction on each step as the pseudo label and set $w_i=1$ only if the pseudo label is consistent with the final answer's correctness.
This dynamic filtering allows the model to avoid noisy samples and focus on the most representative steps of correct and incorrect solutions. 
Thus, by enlarging the score gap between correct and incorrect steps, SPRM can progressively learn the process evaluation ability with only final annotations.

\subsection{Inference with Reflective Generative Form}
In the inference stage, our Reflective Generative Form is naturally suitable for TTS where the SPRM can provide guidance for selecting the high-quality reasoning trajectory from the policy model.
The total inference process divides into three steps(shown in Fig.\ref{fig:img_framework}(b)):
(1) For the given question, the policy model first samples $k$ thinking processes as the candidate reasoning trajectories: $think_1,think_2, \dots, think_k$.
(2) The SPRM evaluates the steps in each process and obtains the final score by the geometric mean of corresponding process scores: $S_1,S_2, \dots, S_k$.
(3)  The reasoning trajectory with the highest final score is chosen and guides the policy model to answer the question (Eq.\ref{eq:tts}).

\begin{equation} \label{eq:tts}
\text{answer} = LLM_{\text{answer}}(think_{j}), \ \text{where} \ j = argmax(S_1, S_2, \dots, S_k)
\end{equation}

\section{Experiment}

\subsection{Baseline\&Dataset}
We conduct experiments on the models with three different sizes: MetaStone-S1-1.5B, 7B, and 32B, which are initialized from DeepSeek-R1-Distill-Qwen-1.5B/7B \citep{deepseekr1}, and QWQ-32B \citep{qwq32b} with continual reinforcement training. 
After adding the SPRM head, only 5M/26M/53M extra parameters are introduced for MetaStone-S1-1.5B/7B/32B.
Our training dataset is constructed from multiple publicly available math-related sources, including NuminaMath~\citep{li2024numinamath}, OpenR1-Math-220k, DeepScaleR~\citep{deepscaler2025}, LIMR~\citep{li2025limr}, and OREAL-RL~\citep{lyu2025exploring}. We apply a multi-agent data cleaning framework to ensure data quality, resulting in a final dataset of 40k high-quality examples.
In the inference stage, we set 3 reasoning efforts with $k=2,8,32$ in Eq.\ref{eq:tts}, named MetaStone-S1-low, -medium, and -high.

We evaluate our models on mathematical benchmarks: AIME2024 and AIME2025 \citep{aime} , and out-of-distribution benchmarks: LivecodeBench(240801-250201) \citep{jain2024livecodebench} and C-Eval \citep{huang2023c}.
AIME2024 and AIME2025 are challenging mathematical benchmarks, designed to assess the mathematical reasoning capabilities of LLMs.
LivecodeBench(240801-250201) contains high-quality programming problems from coding competition websites (LeetCode, AtCoder, and CodeForces), designed to assess the code generation and problem-solving abilities of LLMs.
C-Eval is a comprehensive Chinese evaluation benchmark designed to assess the Chinese knowledge and reasoning abilities of LLMs.
We adopt Pass@1 as the evaluation metric. For each problem, the model generates only one final answer, and the Pass@1 score is computed as the proportion of correctly solved problems. To improve the stability of the results, we repeat the evaluation 64 times and report the average accuracy as the final score.

\begin{table}[!t]
    \centering
    \resizebox{0.8\linewidth}{!}{
    \begin{tabular}{@{}l *{4}{c} @{}}
    \toprule
    \multirow{2}{*}{\textbf{Model}} & \multicolumn{2}{c}{\textbf{Mathematical}} & \multicolumn{2}{c}{\textbf{Out-of-Distribution}} \\
    \cmidrule(lr){2-3} \cmidrule(lr){4-5}
    & \textbf{AIME24} & \textbf{AIME25} & \textbf{LiveCodeBench} & \textbf{C-Eval} \\
    \midrule
    \multicolumn{5}{c}{\textit{Small-size Open-Source Models}} \\
    \textbf{DeepScaleR-1.5B-Preview} & 43.1 & 30.0 & - & - \\
    \textbf{R1-Distill-Qwen-1.5B} & 28.9 & 22.8 & 16.9 & 27.1 \\
    \textbf{R1-Distill-Qwen-7B} & 55.5 & - & 37.6 & -\\
    \textbf{R1-Distill-Llama-8B} & 50.4 & - & 39.6 & -\\
    \midrule
    \textbf{Baseline-1.5B} & 39.3 & 29.9 & 22.4 & 41.8 \\
    \textbf{MetaStone-S1-1.5B-low} & 44.0 & 32.6 & 24.2 & 43.6 \\
    \textbf{MetaStone-S1-1.5B-medium} & 53.1 & 35.7 & 26.6 & 43.9 \\
    \textbf{MetaStone-S1-1.5B-high} & 57.9 & 40.4 & 28.1 & 44.1 \\
    \midrule
    \textbf{Baseline-7B} & 54.7 & 41.2 & 39.4 & 51.3 \\
    \textbf{MetaStone-S1-7B-low} & 60.7 & 45.4 & 41.7 & 55.1 \\
    \textbf{MetaStone-S1-7B-medium} & \underline{66.3}  & \underline{48.3} & \underline{44.1} & \underline{57.5} \\
    \textbf{MetaStone-S1-7B-high} & \textbf{70.2}  & \textbf{48.6} & \textbf{44.4} & \textbf{57.8} \\
    \midrule
    \midrule
    \multicolumn{5}{c}{\textit{Large-size Open-Source Models}} \\
    \textbf{s1-32B} & 56.7 & 50.0 & - & - \\
    \textbf{QwQ-32B} & 79.5 & 69.5 & 63.4 & 88.4 \\
    \textbf{R1-Distill-Qwen-32B} & 72.6 & 49.6 & 57.2 & 82.2 \\
    \textbf{GLM-Z1-32B-0414} & 80.8 & 63.6 & 59.1 & - \\
    \textbf{DeepSeek-R1-671B} & 79.8 & 70.0 & \underline{65.9} & \textbf{91.8} \\
    \multicolumn{5}{c}{\textit{Closed-Source Models}} \\
    \textbf{Claude-3.5-Sonnet1022} & 16.0 & 7.4 & 37.2 & 76.7 \\
    \textbf{GPT-4o-0513} & 9.3 & 11.6 & 32.9 & - \\
    \textbf{OpenAI o1-mini} & 63.6 & 50.7 & 53.8 & 68.9 \\
    \textbf{OpenAI o1-1217} & 79.2 & - & 63.4 & - \\
    \textbf{OpenAI o3-mini*} & 79.6 & \textbf{74.8} & \textbf{67.4} & 75.9 \\
    \midrule
    \textbf{Baseline-32B} & 79.9 & 70.5 & 63.4 & 89.4 \\
    \textbf{MetaStone-S1-32B-low} & 82.0 & 72.0 & 63.8 & 89.5 \\
    \textbf{MetaStone-S1-32B-medium} & \underline{84.2} & 73.4 & 64.0 & 89.6 \\
    \textbf{MetaStone-S1-32B-high} & \textbf{85.2} & \underline{73.6} & 64.2 & \underline{89.7} \\
    \bottomrule
    \end{tabular}
    }
    \caption{Comparison of MetaStone-S1 models and other comparable models. * denotes medium level of OpenAI o3-mini. The best and second-best results are shown in \textbf{bold} and \underline{underlined}.}
    \label{tab:main_res}
\end{table}

\subsection{Main Results}
Table~\ref{tab:main_res} summarizes the performance of our MetaStone-S1 models across four representative benchmarks. 
We denote baseline as the only policy model without using the Reflective Generative Form.
Across different model scales, our proposed Reflective Generative Form consistently enhances the baseline, particularly on mathematical reasoning benchmarks. Specifically, compared with the baseline, MetaStone-S1 achieves performance gains of 18.6/15.5/5.3 points on AIME24 and 10.5/7.4/3.1 points on AIME25 for the 1.5B/7B/32B sizes, respectively. For other tasks, our Reflective Generative Form continues to offer stable improvements, yielding gains of 5.7/5.0/0.8 points on LiveCodeBench and 2.3/6.5/0.3 points on C-Eval.

We further compare MetaStone-S1 against both advanced open-source and closed-source models. 
Among the open-source models, we consider DeepScaleR-1.5B-Preview~\citep{deepscaler2025}, DeepSeek-R1-Distill-Qwen (1.5B/7B/32B), DeepSeek-R1-Distill-LLaMA-8B~\citep{deepseekr1}, QwQ-32B~\citep{qwq32b}, GLM-Z1-32B-0414~\citep{glm2024chatglm}, s1-32B~\citep{muennighoff2025s1}, and DeepSeek-R1-671B~\citep{deepseekr1}. 
Among the closed-source models, we include Claude-3.5-Sonnet-1022, GPT-4o-0522, OpenAI o1-mini, OpenAI o1-1217, and OpenAI o3-mini-medium.

At the small scale, MetaStone-S1-1.5B and MetaStone-S1-7B consistently outperform the listed open-source models with comparable or larger parameter sizes. Especially, MetaStone-S1-1.5B-low surpasses DeepScaleR-1.5B-Preview and R1-Distill-Qwen-1.5B on all datasets.
And MetaStone-S1-1.5B-high further outperforms R1-Distill-Qwen-7B and R1-Distill-Llama-8B on AIME24 (57.9\% vs 55.5\%), demonstrating strong efficiency and capability of our lightweight SPRM.
For MetaStone-S1-7B, its low reasoning effort has outperformed R1-Distill-Qwen-7B and R1-Distill-Llama-8B on AIME24 and LiveCodeBench.
And MetaStone-S1-7B-high further gains the improvement of 14.7 points on AIME24 (70.2\% vs 55.5\%) and 4.8 points on LiveCodeBench (44.4\% vs 39.6\%).

At the larger scale, MetaStone-S1-32B-high achieves superior results on mathematical reasoning tasks, outperforming all listed open-source models of comparable or even larger size by +4.4\% on AIME24 (85.2\% vs 80.8\%) and +3.6\% on AIME25 (73.6\% vs 70.0\%). 
For other out-of-distribution tasks, MetaStone-S1-32B-high surpasses other 32B-sized models by +0.8\% on LiveCodeBench (64.2\% vs 63.4\%) and +1.3\% on C-Eval (89.7\% vs 88.4\%).
Compared to closed-source models, medium and high levels of MetaStone-S1-32B outperform Claude-3.5-Sonnet-1022 and GPT-4o-0522, and achieving comparable performance with medium level of OpenAI o3-mini (85.2\% vs 79.6\% on AIME24, 73.6\% vs 74.8\% on AIME25, 64.2\% vs 67.4\% on LiveCodeBench, 89.7\% vs 75.9\% on C-Eval,), highlighting its strong competitiveness in both mathematical and general reasoning tasks.

\begin{figure*}[!b] 
  \centering
  \includegraphics[width=0.5\textwidth]{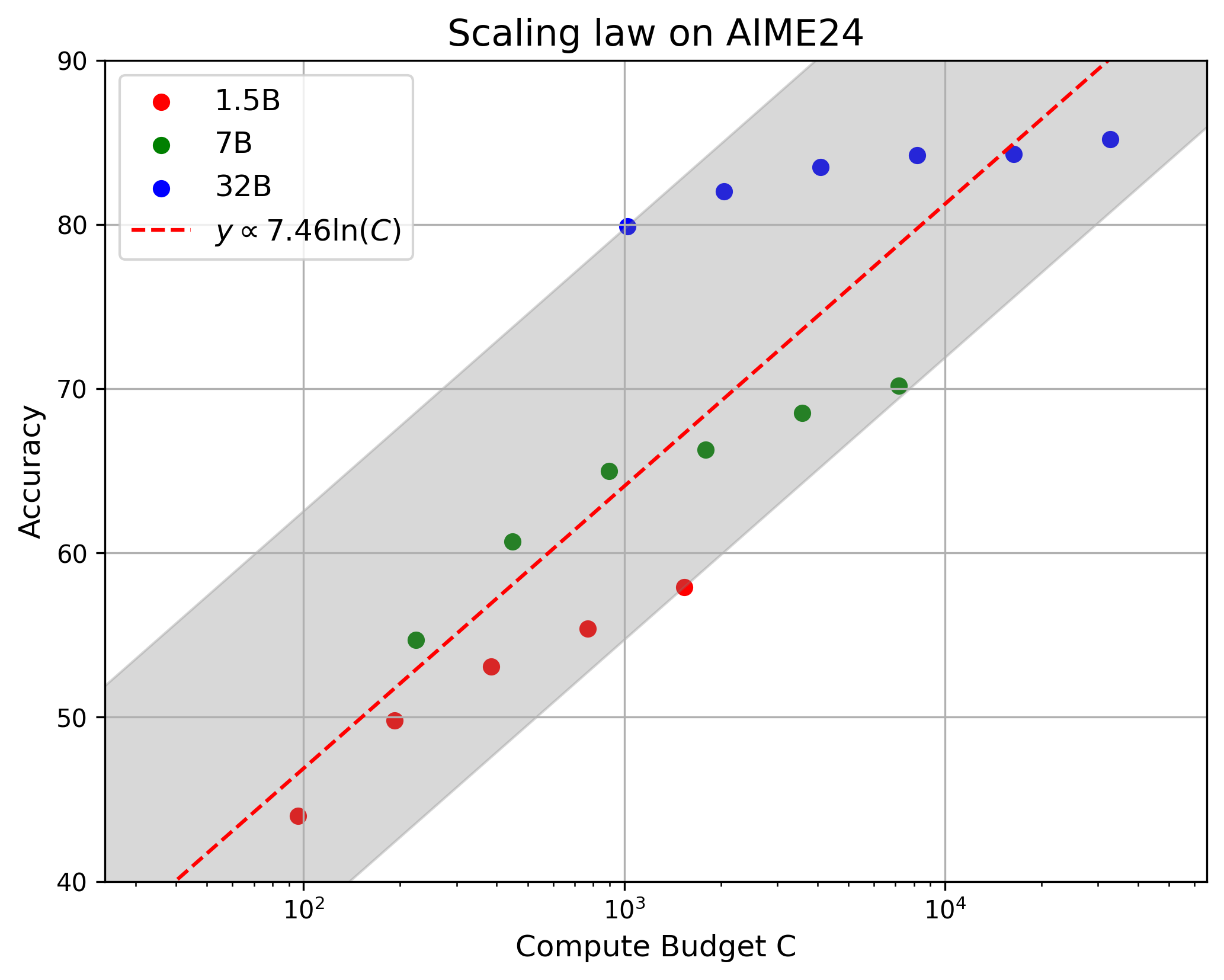}
  \caption{The scaling law of Reflective Generative Models.}
  \label{fig:img_scaling}
\end{figure*}

\subsection{Scaling Law of the Reflective Generative Model}
In Fig.~\ref{fig:img_scaling}, we present the scaling law for reflective generative models, which shows the relationship between the total reasoning computation in TTS and the final performance.
Following \citet{snell2024scaling}, we define the computation budget 
$C$ as the product of the model’s parameter (B) and the total number of reasoning tokens (k): $C=Param_{policy}\times Token_{infer}$.
Notably, when the total reasoning length is scaled to more than 32 times the baseline (e.g., Best-of-64), the performance improves slowly.
Therefore, we mainly focus on TTS results up to Best-of-32 for each model scale. 
We observe that the final performance shows a positive correlation with the logarithm of the computation budget (the specific scaling factor depends on the baseline model architecture).
This indicates that the final performance of our model can be enhanced by exponentially scaling on parameter size or the reasoning length.

\begin{figure*}[!b] 
  \centering
  \includegraphics[width=\textwidth]{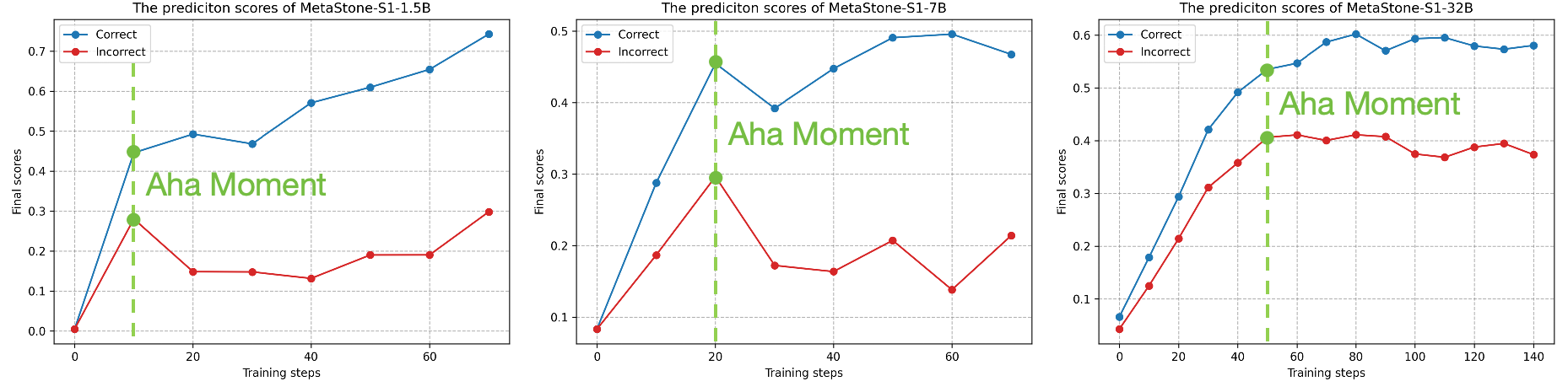}
  \caption{The training process of SPRM. The blue and red curves denote the final score on correct and incorrect reasoning trajectories. The green dashed line indicates the "aha moment".}
  \label{fig:img_score}
\end{figure*}

\begin{figure}[!b]
\centering
\includegraphics[width=0.9\textwidth]{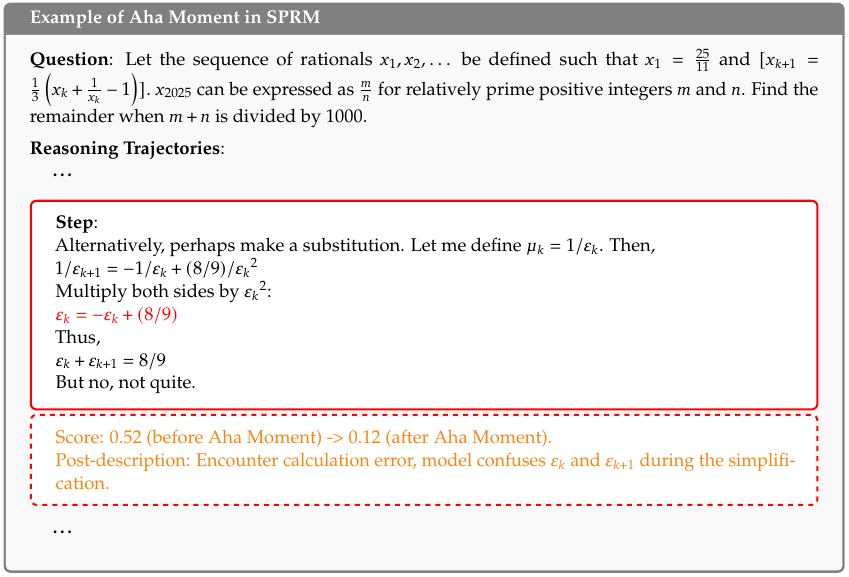}
\caption{Comparison of SPRM's predictions before and after the aha moment. Only key steps are listed. The error steps are marked in red.}
\label{fig:aha_example}
\end{figure}

\subsection{Aha Moment of the Reflective Generative Model}

In Fig.~\ref{fig:img_score}, we present the final evaluation scores from MetaStone-S1 for both correct and incorrect reasoning trajectories throughout the training process.
During the initial phase of training, the optimization trajectories for all samples follow a similar trend, indicating that the model has not yet learned to distinguish between correct and incorrect reasoning trajectories.
However, after a certain number of training steps (e.g., around 10/20/50 steps, 1280/2560/6400 samples for MetaStone-S1-1.5B/7B/32B, respectively), we observe a distinct "aha moment" point where the optimization trends of different reasoning trajectories begin to diverge.
This suggests that the model is starting to judge the correctness based on the reasoning contents.
With this aha moment, MetaStone-S1 can progressively refine its SPRM head through our proposed self-supervised SPRLoss, leading to a clear score gap between correct and incorrect reasoning trajectories, and enabling effective TTS.

Specially, we show an example in Fig.\ref{fig:aha_example}. We fix the reasoning trajectory and use MetaStone-S1 before and after the aha moment for scoring. In this case, the model mistakenly confuses $\varepsilon_k$ and $\varepsilon_{k+1}$, resulting in an incorrect solution. Our model fails to recognize the error before the aha moment, while the model after the aha moment can correctly discriminate it.

\begin{table}[t]
    \centering
    \resizebox{\linewidth}{!}{
    \begin{tabular}{@{}l l @{}*{5}{c} @{}}
    \toprule
    \centering\textbf{Model} & \centering\textbf{Reward Model} & \textbf{Extra Params} & \textbf{AIME24} & \textbf{AIME25} & \textbf{LiveCodeBench} \\
    \midrule
    \multirow{3}{*}{\centering MetaStone-S1-1.5B-high}
     & Qwen2.5-Math-RM & 72B & 55.8 & 35.3 & 26.8 \\
     & Qwen2.5-Math-PRM & 72B & 56.7 & 40.0 & 26.8 \\
     & SPRM & \textbf{5M} & \textbf{57.9} & \textbf{40.4} & \textbf{28.1} \\
    \midrule
    \multirow{3}{*}{\centering MetaStone-S1-7B-high}
     & Qwen2.5-Math-RM & 72B & 63.5 & 46.7 & 42.3 \\
     & Qwen2.5-Math-PRM & 72B  & 68.8 & 48.3 & 42.8 \\
     & SPRM & \textbf{26M} & \textbf{70.2} & \textbf{48.6} &  \textbf{44.4} \\
    \bottomrule
    \end{tabular}
    }
    \caption{Comparison of SPRM and other PRM models.}
    \label{tab:sprm_model}
\end{table}

\begin{figure*}[t] 
  \centering
  \includegraphics[width=0.9\textwidth]{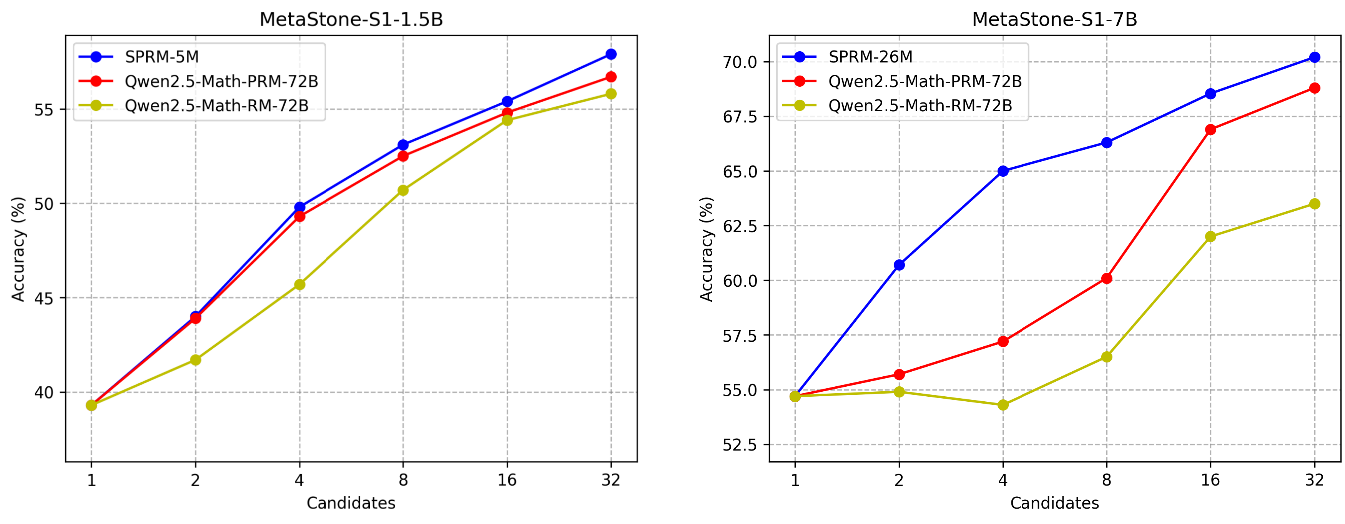}
  \caption{Evaluation of varying numbers of candidate reasoning trajectories on AIME24.}
  \label{fig:img_bon}
\end{figure*}

\begin{figure}[!p]
\centering
\includegraphics[width=0.9\textwidth]{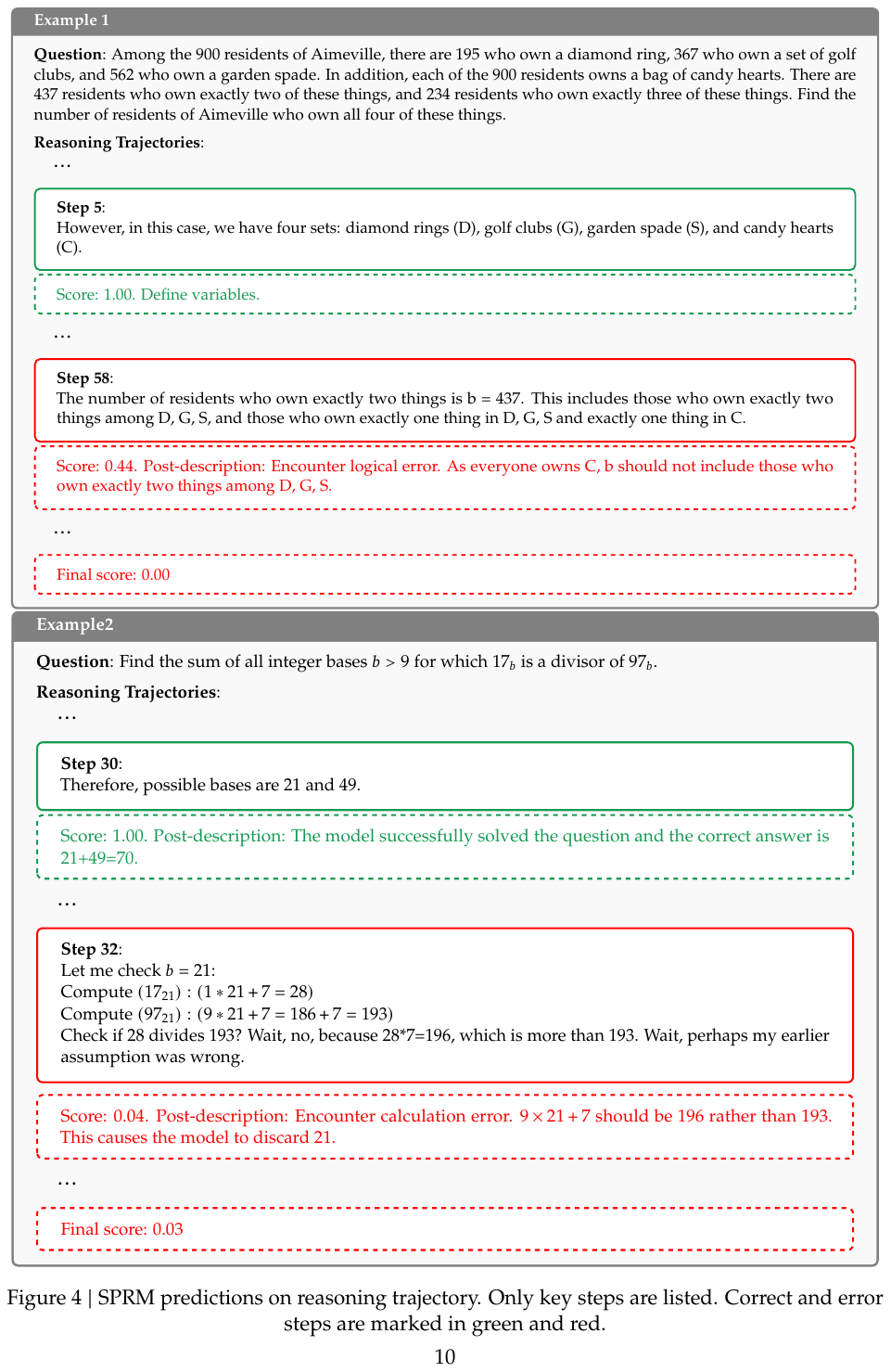}
\caption{SPRM's predictions on reasoning trajectories. Only key steps are listed. Correct and error steps are marked in green and red.}
\label{fig:sprm_example}
\end{figure}

\subsection{Ablation Study}

\paragraph{Effectiveness of SPRM.}  
As shown in Table.\ref{tab:sprm_model}, our SPRM, with only around 26M extra parameters, achieves higher performance compared with 72B ORM and PRM. 
This demonstrates that sharing parameters between the reward model and policy model is able to generate high-quality guidance without the requirement of additional large-scale reward models.
To further analyze the efficacy of different reward models, we compare their performance under different numbers of candidate reasoning trajectories $k$ in Fig.\ref{fig:img_bon}. 
Across different $k$ and model sizes, SPRM consistently outperforms other methods, indicating its strong ability to distinguish between high and low quality reasoning trajectories.

Moreover, we evaluate the generalization ability of SPRM on tasks from the other domain (LiveCodeBench).
Without any task-specific fine-tuning or in-domain data, our SPRM still achieves superior performance compared to separate reward models.
This demonstrates SPRM's strong zero-shot generalization capability, suggesting that it can capture domain-agnostic patterns to evaluate the reasoning trajectories.

Fig.\ref{fig:sprm_example} shows the visualization of step-wise evaluation scores from SPRM. It can be observed that SPRM effectively identifies low-quality processes generated by the policy model, including logical error (e.g. the misunderstanding of b in step 58 of example 1) and calculation error (e.g., the incorrect computation 9 × 21 + 7 = 193 in step 32 of example 2). SPRM assigns low scores to these low-quality steps. Since SPRM only outputs process scores, we additionally provide post-descriptions within the dashed boxes for better clarity.

\begin{figure*}[!t] 
  \centering
  \includegraphics[width=0.85\textwidth]{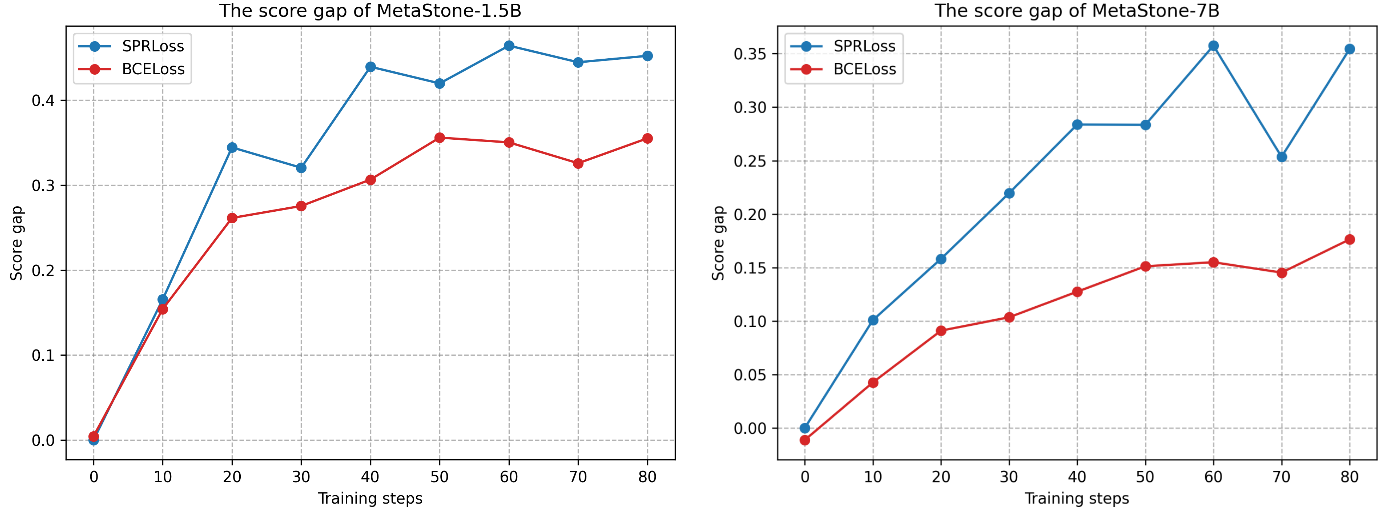}
  \caption{The prediction score gap between correct and incorrect solutions. The blue curve shows the SPRLoss. The red curve shows the BCELoss.}
  \label{fig:img_sprloss}
\end{figure*}

\paragraph{Effectiveness of self-supervised optimization.}  
We evaluate the effectiveness of SPRLoss in Table.\ref{tab:sprm_loss}.
Compared with using the final answer correctness as process-level supervision for PRM training,
our proposed self-supervised optimization method achieves larger performance gains on both 1.5B and 7B models.
Furthermore, Fig.\ref{fig:img_sprloss} shows the prediction score gap between correct and incorrect solutions.
Compared to the BCELoss, SPRLoss demonstrates stronger discriminative capability with a larger score gap.
This indicates that treating final answer correctness as process-level labels introduces substantial label noise, which harms the optimization.
In contrast, SPRLoss leverages self-supervised signals to reduce the impact of noisy supervision, leading to more stable and accurate training.

\begin{table}[t]
    \centering
    \begin{tabular}{@{}l l *{2}{c} @{}}
    \toprule
    \centering\textbf{Model} & \centering\textbf{Loss} & \textbf{AIME24} & \textbf{LiveCodeBench} \\
    \midrule
    \multirow{2}{*}{\centering MetaStone-S1-1.5B-high}
     & BCELoss & 56.7 &  27.9 \\
     & SPRLoss & \textbf{57.9} & \textbf{28.1} \\
    \midrule
    \multirow{2}{*}{\centering MetaStone-S1-7B-high}
     & BCELoss & 69.1 & 43.9 \\
     & SPRLoss & \textbf{70.2} & \textbf{44.4} \\
    \bottomrule
    \end{tabular}
    \caption{Evaluation on SPRLoss.}
    \label{tab:sprm_loss}
\end{table}

\begin{table}[!h]
    \centering
    \begin{tabular}{@{}l *{5}{c} @{}}
    \toprule
    \textbf{Searching Tokens(k)} & 0 & 40 & 80 & 120 & 160 \\
    \midrule
    \textbf{Accuracy(\%)} & 39.3 & 48.8 & 50.0 & 51.7 & 52.8 \\
    \bottomrule
    \end{tabular}
    \caption{Performance of MetaStone-S1-1.5B with MCTS on AIME24.}
    \label{tab:mcts}
\end{table}

\subsection{Extend on MCTS}

Since SPRM produces process scores for each step, our reflective generative models can be naturally used in step-level search-based TTS methods such as Monte Carlo Tree Search (MCTS). 
In our setup, instead of performing full simulations to the end of a reasoning trajectory, we directly use SPRM to estimate the value of each node during the search. This enables a more efficient evaluation at each expansion step. 
In the expanding stage, we expand 4 children for the selected node and generate 1024 tokens in each child node.
To balance the computation cost, we set the maximum number of tokens in the MCTS process from 0 (without MCTS) to 160k for each question.

Table.\ref{tab:mcts} presents the performance of MCTS on the AIME24 dataset. 
By increasing the maximum number of searching tokens, the performance improved from 39.3 to 52.8, demonstrating the effectiveness of SPRM in providing high-quality step-level guidance in the reasoning stage.
However, the performance of MCTS is still lower than the results under the Best-of-N strategy reported in Table~\ref{tab:main_res}. This is primarily due to the computational overhead inherent in tree-based search methods, which leads to incomplete searching in our setting. Nonetheless, the observed gains over the baseline validate the potential of our reflective generative models for integration with more advanced search-based inference methods.

\section{Conclusion}
In this work, we propose a novel Reflective Generative Form, which enables a single LLM to both generate and select high-quality reasoning trajectories for Test-Time Scaling (TTS). Based on this form, we introduce our first reflective generative model, MetaStone-S1, which achieves comparable performance to the OpenAI o3-mini series.
Specifically, we design a unified interface that integrates the policy model and process reward model (PRM) within a single network, resulting in low parameter overhead and efficient TTS inference. 
To further reduce the dependence on costly process-level annotations for PRM training, we present a self-supervised process reward model (SPRM) that learns process-level evaluation using only final answer annotations.
MetaStone-S1, with 32B parameters, achieves strong performance across mathematics, coding, and Chinese reasoning benchmarks, outperforming a series of open-source and closed-source models. In addition, our experimental analyses of the aha moment and scaling law further demonstrate the effectiveness of our Reflective Generative Form.
In future work, we plan to explore its capability for more efficient step-level search-based TTS for real-time reasoning enhancement.


\end{document}